\documentclass{article}

\usepackage{arxiv}

\usepackage[utf8]{inputenc}
\usepackage[T1]{fontenc}
\usepackage{hyperref}
\usepackage{url}
\usepackage{booktabs}
\usepackage{amsmath,amssymb,amsfonts}
\usepackage{microtype}
\usepackage{graphicx}
\usepackage[numbers,sort&compress]{natbib}
\usepackage{xcolor}
\usepackage{multirow}
\usepackage{float}
\usepackage{caption}
\hypersetup{colorlinks=true,linkcolor=blue!55!black,citecolor=blue!55!black,urlcolor=blue!55!black}

\newcommand{\jev}{Jev}
\newcommand{\qwen}{Qwen3.8-27B}
\newcommand{\mad}{\mathrm{MAD}}
\newcommand{\nf}{\mathrm{NF}}
\newcommand{\ci}[2]{\begin{tabular}[t]{@{}c@{}}#1\\[-1.5pt]{\scriptsize$[#2]$}\end{tabular}}

\title{Beyond Calibration: Do a Typed-Decision Model's Probabilities Obey the Probability Axioms?}

\author{Keyi Li \qquad Yihao He \qquad Quanyi Li \\
	Software College, Northeastern University \\
	Shenyang, China
}

\renewcommand{\shorttitle}{Beyond Calibration: Probability Axioms for Typed Decisions}

\hypersetup{
pdftitle={Beyond Calibration: Do a Typed-Decision Model's Probabilities Obey the Probability Axioms?},
pdfauthor={Keyi Li, Yihao He, Quanyi Li},
pdfkeywords={probabilistic coherence, calibration, negation, typed-decision models, LLM probability elicitation},
}

\begin{document}
\maketitle

\begin{abstract}
Typed-decision models such as TypeSafe's \jev{} answer a declared yes/no or multiple-choice question about a state with a probability instead of text, and their evaluations report accuracy and calibration. Neither requires that the probabilities a model gives to logically related questions fit together item by item, which is what a system that acts on those probabilities needs. We test this property, coherence, with a battery of logically linked questions that needs no labels. For 160 items from ChaosNLI and PubMedQA, each with three mutually exclusive labels, we ask whether the label is $X$, whether it is not $X$, whether it is one of the other two labels, and which label applies. On 480 negation pairs, \jev{}'s probabilities for ``the label is $X$'' and ``the label is not $X$'' miss summing to one by 0.064 on average (95\% CI 0.055 to 0.072). \qwen{}, run from its official BF16 weights, misses by 0.293 with first-token probabilities and by 0.122 with verbalized probabilities. The gap to the first-token readout persists on pairs where both systems give similar probabilities, without double-negation labels, and after averaging \jev{}'s repeated calls. \jev{} is not coherent either: its violations are about five times its repeat noise, and it over-endorses statements about single labels, so that its three single-label probabilities sum to 1.14 on average. The two systems also fail differently. \qwen{}'s first-token readout under-endorses the complement of a label whether or not the question contains ``not'', rejecting both a statement and its negation in 196 of 480 pairs, and it does not become more coherent where it is more confident, whereas \jev{}'s violations concentrate where its answer is uncertain. Because the checks need no labels, they expose biases that appear only when question forms are compared, and inconsistencies within items.
\end{abstract}

\keywords{probabilistic coherence \and calibration \and negation \and typed-decision models \and probability elicitation}

\section{Introduction}\label{sec:intro}

Automated pipelines increasingly act on model probabilities. A threshold on $P(\text{yes})$ decides whether a claim is flagged, an expected-cost rule decides whether to escalate to a human, and several probabilities about the same case are combined into one decision. A recent class of \emph{typed-decision} models is built for this use. TypeSafe's \jev{}, described by its developer as a ``System One'' model trained for calibrated decisions, reads an unstructured state and answers declared questions: a \emph{noul} question returns the probability of ``yes'', and a \emph{choice} question returns a distribution over the declared options~\citep{typesafe2026jev,openrouter2026decisions}. Because the probability is the product, evaluations of \jev{} and similar models measure accuracy, proper scores, and calibration~\citep{jevals2026,rafe2026crash,cheng2026thisthat}.

These measures leave a gap. Calibration compares stated probabilities with outcome frequencies, averaged over many cases~\citep{guo2017calibration}, and proper scores reward it together with sharpness~\citep{gneiting2007scoring}. A system can still assign probability 0.7 to ``the claim is supported'' and 0.6 to ``the claim is not supported'' for the same case. Such an assignment violates the probability axioms~\citep{kolmogorov1956foundations}, and a decision maker who acts on it can be led into a sure loss~\citep{definetti1974theory}. In a pipeline the failure is concrete: the action depends on whether the question was phrased positively or negatively, or on whether the options were asked about one at a time or together. Coherence, the agreement of probabilities across logically related questions, is therefore a property that users of native probabilities rely on. It is related to calibration, since a coherent forecaster expects to be calibrated~\citep{dawid1982calibrated} and incoherent forecasts are dominated under proper scoring rules~\citep{predd2009coherence}, but calibration measured on one question form does not establish agreement across forms, a distinction recently drawn for the confidence estimates of LLMs~\citep{matta2026rethinking}.

For large language models, incoherence is well documented. Their probability judgments violate identities that hold for any probability measure~\citep{zhu2024incoherent}, their forecasts fail negation and monotonicity checks~\citep{fluri2023consistency,paleka2024forecasters}, their answers change under negation and paraphrase~\citep{kassner2020negated,elazar2021consistency,jang2022becel}, and different ways of eliciting a probability from the same model disagree~\citep{tian2023justask,wang2024myanswer}. For typed-decision models the question is newer. Benchmark evaluations of \jev{} score each question in a single form~\citep{jevalsmethod2026}, so an inconsistency between forms does not show up in them. The developer states that such invariants are not guaranteed~\citep{typesafe2026jaggedness}. Community audits released in the model's first week measured violations of negation complementarity and of agreement between yes/no and choice questions~\citep{jujumilk2026audit,yodablocks2026orderby,li2026jevjudge}, and one found \jev{} more coherent than untuned open-LLM readouts on an aggregate score~\citep{jevify2026} (Section~\ref{sec:related}). They leave open how large each violation is relative to repeat noise, whether further identities fail, and whether a comparison with an LLM survives differences in confidence, readout, and double negation.

We test coherence directly. For each item we fix a state and an explicit partition into three mutually exclusive, exhaustive labels, and we ask ten questions: whether the label is $X_k$, whether it is not $X_k$, whether it is one of the other two labels, and which label applies. Probability theory fixes the relations among the ten answers, so every item provides several checks that need no gold label (Section~\ref{sec:battery}). We compare \jev{} with the official release of \qwen{}, an open-weight LLM run in BF16, on identical questions under two probability readouts, the first-token probability of the answer and a probability stated in JSON (Section~\ref{sec:setup}).

The results support three claims.
\begin{itemize}
\item \textbf{\jev{} violates negation complementarity less than either LLM readout, and far less than the first-token readout.} Its mean absolute violation of $s_k+n_k=1$ is 0.064, against 0.293 for \qwen{}'s first-token readout, a 4.6-fold difference, and 0.122 for its verbalized readout. The difference holds on pairs where both systems give similar probabilities, without the two labels whose negation is a double negative, and after averaging \jev{}'s repeated calls (Sections~\ref{sec:neg} and~\ref{sec:robust}). The verbalized readout deviates by more than 0.1 less often than \jev{} does, in 11.9\% of pairs against 19.0\%, but by much more when it does.
\item \textbf{\jev{} is not coherent.} Its violation is about five times its repeat-noise floor. A label-free decomposition locates the systematic part: \jev{} gives single-label statements an average probability of 0.380 where coherence requires 1/3, which also makes its three single-label probabilities sum to 1.14. Its yes/no and choice probabilities for the same label differ by 0.09 (Sections~\ref{sec:beyond} and~\ref{sec:forms}).
\item \textbf{The failure signatures differ.} \qwen{}'s first-token readout under-endorses the complement of a label whether it is phrased with ``not'' or as a disjunction, answers ``no'' to both a statement and its negation in 196 of 480 pairs, and violates the identity no less where it is confident than where it is not. Its verbalized readout errs in the opposite direction. \jev{}'s violations concentrate where its answer is uncertain (Section~\ref{sec:where}).
\end{itemize}

We contribute an identity-by-identity coherence battery for typed decisions with a measured repeat-noise floor and controls for confidence, readout, and double negation, and we use it to compare a typed-decision model with two readouts of an LLM on identical questions. In an exploratory analysis we add a label-free decomposition that locates each system's systematic deviations in specific question forms. The direction of our main result agrees with earlier community audits~\citep{jevify2026}. Of two patterns that support theory predicts for human judgment, subadditivity over a partition appears in \jev{} but not in \qwen{}, and neither system shows the unpacking effect.

\section{Background and related work}\label{sec:related}

\paragraph{Typed-decision models.}
\jev{} is a closed model, served through an API, that answers typed questions about a state in a single call~\citep{typesafe2026jev}. Open imitations and related typed-decision models followed its release~\citep{laya2026,cheng2026thisthat,ren2026openjev}. Evaluations and applications so far concern accuracy, calibration, and confidence-based escalation~\citep{jevals2026,rafe2026crash,li2026jevjudge,wu2026reflex,deng2026scientific}. \citet{sun2026typesafe} show that a constrained decision head can follow the option name rather than the rubric attached to it, which also concerns how the question is worded. The independent Jevals benchmark scores native \jev{} probabilities against verbalized LLM probabilities, which it notes tend to cluster on round values~\citep{jevalsmethod2026}. Its measures are proper scores, calibration, and flip rates under repeats and option reordering, none of which asks logically related questions about the same state.

\paragraph{Coherence of probability judgments.}
Coherence is the classical requirement for a set of probability assessments: violating the axioms exposes the assessor to a sure loss~\citep{definetti1974theory,kolmogorov1956foundations}. Its relation to calibration has a long history in the forecasting literature~\citep{dawid1982calibrated,seidenfeld1985calibration,predd2009coherence}. Human judgments are incoherent in systematic ways. Support theory~\citep{tversky1994support,rottenstreich1997unpacking} predicts that judgments of a hypothesis and its complement sum to about one, that judgments of three or more exhaustive hypotheses sum to more than one, and that describing a hypothesis as an explicit disjunction of its parts raises its judged probability. The conjunction fallacy is another classic violation~\citep{tversky1983extensional}. Noise-based accounts explain why some identities hold on average while others fail~\citep{costello2014surprisingly,zhu2020bayesiansampler}. We borrow these identities and test two of the support-theory patterns on both systems.

\paragraph{Consistency of language models.}
Consistency checks for language models cover paraphrase~\citep{elazar2021consistency}, negation, symmetry, and transitivity~\citep{jang2022becel,jang2023chatgpt}, negated probes~\citep{kassner2020negated,hosseini2021understanding,truong2023naysayers,garciaferrero2023thisisnot}, and systems of beliefs~\citep{kassner2021beliefbank,mitchell2022concord,hase2023beliefs}. Negation consistency, $p(x^{+}) \approx 1-p(x^{-})$, also serves as an unsupervised objective for probes that recover latent knowledge from hidden states~\citep{burns2022ccs}. For probabilities specifically, consistency checks detect errors of forecasters without ground truth~\citep{fluri2023consistency,paleka2024forecasters}, LLM probability judgments are incoherent in ways reminiscent of human judgments~\citep{zhu2024incoherent}, and fact-checking answers to queries with negation, conjunction, and disjunction are often logically inconsistent~\citep{ghosh2024logical}. Direct tests of probabilistic coherence cover complementarity and monotonicity under prompted and logit readouts~\citep{freedman2025rational}, logical constraints on token probabilities across model families~\citep{richardson2026modellog}, Dutch-book arbitrage over verbalized forecasts~\citep{andrews2026dutch}, and, earlier, the coherence of a non-chat model's token probabilities~\citep{betz2023coherence}. Results depend on the model: GPT-5 satisfies binary complementarity almost perfectly in a comparison with human judges~\citep{imannezhad2026divergent}, and interventions that reduce calibration error can leave structural violations unchanged~\citep{matta2026rethinking}. Training objectives that enforce logical constraints have been proposed~\citep{calanzone2024logically}, as have checks for self-contradiction and for agreement between generating and validating an answer~\citep{mundler2023selfcontradictory,li2023generatorvalidator}. Behavioral test suites such as CheckList~\citep{ribeiro2020checklist} use the same logic of invariance tests.

\paragraph{Coherence audits of \jev{}.}
Several audits of \jev{}'s coherence appeared as code repositories within days of its release. The closest to our study, Jevify, builds 4{,}749 families of related questions, including negation pairs and yes/no versions of choice options, and reports a de Finetti sure loss of 0.081 for \jev{} against 0.141 to 0.205 for untuned open-LLM readouts~\citep{jevify2026}. Other audits report complements that sum to between 0.71 and 1.42 and yes/no and two-option choice answers that differ by 0.125 on average~\citep{jujumilk2026audit}, negation violations of 0.016 on topic classification that are no larger than those produced by a paraphrase~\citep{yodablocks2026orderby}, and, on 48 judging examples, a complement residual of 0.045 and a choice-versus-yes/no gap of 0.055~\citep{li2026jevjudge}. \citet{rafe2026crash} show that \jev{}'s probabilities lie on a 0.01 grid and never reached 0 in 2.4 million answers. Negation complementarity and cross-format agreement have therefore been measured before, in aggregate or separately. Our study adds a battery that reports each identity against a measured noise floor, repeats the comparison with an LLM under controls for confidence, readout, and double negation, and adds checks of additivity, description invariance, and monotonicity.

\paragraph{Eliciting probabilities from LLMs.}
An LLM has no single probability for an answer. Token probabilities of an answer can be informative about correctness~\citep{kadavath2022know,jiang2021know,desai2020calibration}, verbalized probabilities are an alternative for instruction-tuned models~\citep{tian2023justask,xiong2024express}, and the two can disagree. The first-token probability of an answer option need not match the answer the model would write~\citep{wang2024myanswer,wang2024lookattext}, probability mass is split across surface forms~\citep{holtzman2021surface}, and label tokens carry prior biases~\citep{zhao2021calibrate}. These effects are why we report two readouts of \qwen{} and check whether the comparison with \jev{} points in the same direction under both.

\section{A coherence battery for typed decisions}\label{sec:battery}

\paragraph{Questions.}
Each item consists of a state $x$ and three labels $X_0,X_1,X_2$ that are mutually exclusive and exhaustive. Every question repeats the same definition paragraph, which defines the three labels and states that exactly one applies. For each label $k$ we ask three yes/no questions and, once per item, one choice question:
\begin{align*}
S_k &: \text{``Is it true that the label is } X_k\text{?''} & N_k &: \text{``Is it true that the label is not } X_k\text{?''}\\
D_k &: \text{``Is it true that the label is either } X_i \text{ or } X_j\text{?''} & \mathit{CH} &: \text{``Which label applies?''}
\end{align*}
where $\{i,j\}=\{0,1,2\}\setminus\{k\}$. $S_k$ and $N_k$ differ only by the word ``not''. We use ``Is it true that the label is not $X_k$?'' rather than ``Is the label not $X_k$?'' because English negative questions can be read as leading questions. $N_k$ and $D_k$ describe the same event, the complement of $X_k$, once through negation and once as an explicit disjunction. Appendix~\ref{app:prompts} gives the full text.

\paragraph{Identities.}
Write $s_k$, $n_k$, $d_k$ for the probabilities of ``yes'' to $S_k$, $N_k$, $D_k$, and $c_k$ for the probability of $X_k$ in the choice question. If all ten answers come from one probability measure $P$ over the labels, then $s_k=c_k=P(X_k)$ and $n_k=d_k=1-P(X_k)$. The battery checks the consequences:
\begin{equation}
\begin{aligned}
&\text{negation complement:} && s_k+n_k=1, &\qquad &\text{partition additivity:} && \textstyle\sum_k s_k=1,\\
&\text{disjunctive complement:} && s_k+d_k=1, & &\text{disjunction additivity:} && d_k=s_i+s_j,\\
&\text{description invariance:} && d_k=n_k, & &\text{monotonicity:} && s_m\le d_k,\ s_m\le n_k\ (m\neq k),\\
&\text{cross-format consistency:} && s_k=c_k. & & & &
\end{aligned}
\label{eq:identities}
\end{equation}
A further invariance concerns the request rather than the question. \jev{} accepts several questions in one request, so we also ask all ten questions of an item together and compare the answers with those of separate requests. None of these checks uses the gold label. \jev{}'s developer notes that separately asked questions need not satisfy such identities, and that a choice question is relative to its options while a yes/no question is absolute~\citep{typesafe2026jaggedness}. Our questions state the partition explicitly, so each identity follows from the question text alone, and the battery measures how far the answers depart from it.

\paragraph{Measures.}
The primary measure is the mean absolute violation of negation complementarity,
\begin{equation}
\mad_{\text{neg}}=\frac{1}{|I|}\sum_{i\in I}\frac{1}{|K_i|}\sum_{k\in K_i}\bigl|s_{ik}+n_{ik}-1\bigr|,
\label{eq:mad}
\end{equation}
where $K_i$ holds the labels of item $i$ whose pair $(S_k,N_k)$ was answered by both systems, so every item has equal weight and both systems are compared on the same pairs. The signed mean of $s_k+n_k-1$ gives the direction: positive values mean that the two answers of a pair together give too much probability to ``yes''. The same-side rate is the share of pairs answered on the same side of 0.5, $(s_k-0.5)(n_k-0.5)>0$, which a threshold decision turns into accepting or rejecting both a statement and its negation. The other identities in Eq.~\eqref{eq:identities} are measured by item means of the absolute and signed deviations, and monotonicity by the mean of $\max(0,s_m-q)$ with $q\in\{d_k,n_k\}$ the consequent, together with the share of checks violated by more than 0.05.

\paragraph{Confounds.}
Four confounds could make one system look more coherent than another without being so. We address each with a dedicated check.
\emph{Confidence.} On the probability scale, a system that answers near 0 or 1 has little room for $|s_k+n_k-1|$, so a more confident or better-memorizing system looks more coherent. We therefore compare the systems on pairs where they give similar probabilities to $S_k$, $|s^{J}_k-s^{Q}_k|\le 0.10$. We also report the logit-scale violation $|\mathrm{logit}\,s_k+\mathrm{logit}\,n_k|$ with probabilities clipped to $[0.01,0.99]$. It weights deviations near 0 and 1 more heavily and depends on the clipping bound, so it serves as a secondary check.
\emph{Readout.} A difference could reflect how probabilities are extracted from the LLM rather than the LLM itself, so we compare \jev{} with both readouts.
\emph{Polarity.} Two of our labels, \emph{contradiction} in NLI and \emph{negative} in PubMedQA, denote falsity, so ``not contradiction'' is semantically a double negative that a model may read as the opposite label. We repeat the comparison without these labels.
\emph{Repeat noise.} Identical requests to \jev{} return slightly different answers, and noise alone inflates $|s_k+n_k-1|$. We ask every \jev{} question twice and estimate a noise floor
\begin{equation}
\nf=\frac{1}{|I|}\sum_{i}\frac{1}{|K_i|}\sum_{k}\frac{\bigl|(s^{(0)}_{ik}-s^{(1)}_{ik})+(n^{(0)}_{ik}-n^{(1)}_{ik})\bigr|}{\sqrt{2}},
\label{eq:nf}
\end{equation}
the violation that a coherent system with the same repeat noise would show. For independent Gaussian noise the numerator has $\sqrt{2}$ times the standard deviation of the noise part of $s_k+n_k-1$, so the correction is exact in that case and approximate for the discrete noise we observe. We also compare after averaging the two repeats, and we measure the noise floor of the LLM readout in the same way on repeats of 20 items (Section~\ref{sec:neg}).

\section{Experimental setup}\label{sec:setup}

\paragraph{Materials.}
The NLI domain uses 100 items drawn at random from the 1{,}599 ChaosNLI items built on MNLI~\citep{nie2020chaosnli,williams2018mnli}. Each ChaosNLI item carries 100 human labels, and we use their entropy to measure human disagreement. The median entropy of the sampled items is 1.11 bits. The state is the premise and hypothesis, and the labels are entailment, neutral, and contradiction. The PubMedQA domain uses 60 expert-labelled items~\citep{jin2019pubmedqa}, 20 per final decision, with the research question and context passages as the state and the conclusion paragraph withheld. We rename the answers yes, no, and maybe to \emph{affirmative}, \emph{negative}, and \emph{inconclusive} to avoid questions such as ``is the label not no?''. Nineteen of the 60 PubMedQA items belong to the public Jevals suite~\citep{jevals2026}, on which \jev{} may have been evaluated or tuned, and we report them separately. Items were sampled with fixed seeds, and the option order of the choice question cycles through the six permutations within each domain.

\paragraph{Systems.}
We query \jev{} (\texttt{typesafe/jev-1.13}) through OpenRouter's decisions endpoint. All 3{,}360 responses report the same version, \texttt{jev-1.13-20260917}. \jev{} returns probabilities with two decimals. Each question is sent in its own request, and all requests are repeated once after the first pass is complete, a few minutes later. For the bundling check, each item is also sent once as a single request containing its ten questions under opaque identifiers in shuffled order, so the identifiers do not reveal the logical structure. The comparison LLM is \qwen{}~\citep{qwen38,qwen38_27b}, run from the official weights in BF16 without quantization or fine-tuning. The weights come from a mirror of the official repository, and every file the model loads is identical to the Hugging Face release. We serve the model with vLLM 0.19.1~\citep{kwon2023vllm} on two RTX A6000 GPUs and disable thinking. Both readouts use the same state, definitions, and question text as \jev{}, preceded by a short system prompt (Appendix~\ref{app:prompts}). The \emph{first-token} readout applies the model's chat template, scores the next token, and sums the probability of the tokens among the 100 most likely that equal ``yes'' or ``no'' after whitespace stripping and lowercasing, or the option letters for the choice question, then renormalizes over the answer labels. The unnormalized total is logged as the \emph{mass} of the readout. The \emph{verbalized} readout asks the model to reply with JSON probabilities for ``yes'' and ``no'' and normalizes them. Both readouts are rounded to two decimals to match \jev{}'s resolution. The verbalized readout was collected for the 960 $S_k$ and $N_k$ questions only.

\paragraph{Protocol and statistics.}
We collected 3{,}360 \jev{} calls, 1{,}720 first-token calls (including a repeat of the $S_k$ and $N_k$ questions for 20 items), and 960 verbalized calls. Every call returned a parseable answer. The \jev{} calls cost US\$0.076 in total. Confidence intervals for means are percentile intervals from a paired item bootstrap~\citep{efron1993bootstrap} stratified by domain, with 5{,}000 resamples and seed 0, and the two systems share the resampled items. Intervals for correlations resample pairs or items, as stated where they appear, and whenever items are resampled the pairs of an item stay together. A pair enters a comparison only if both systems answered both of its questions, and for the primary difference we also run a sign-flip permutation test with 10{,}000 permutations. Re-running the analysis from the raw logs reproduces its output exactly. Analyses marked as exploratory were added after the main analysis and are listed in Appendix~\ref{app:posthoc}.

\section{Results}\label{sec:results}

\subsection{Negation complementarity}\label{sec:neg}

\begin{figure}[t]
\centering
\includegraphics[width=\linewidth]{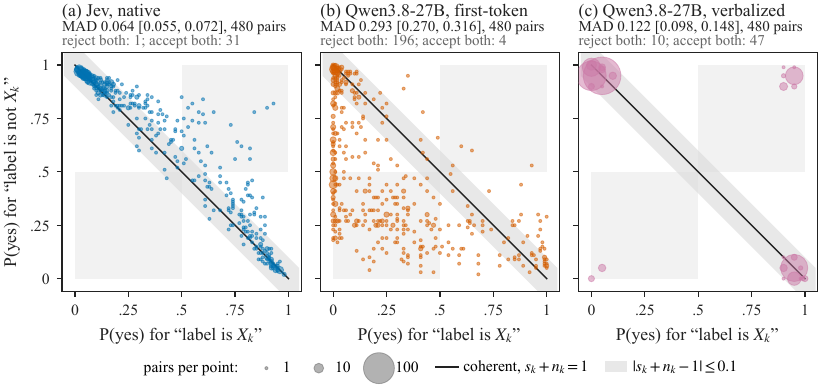}
\caption{Negation pairs. Each point is one pair $(s_k,n_k)$ for an item and label, and point area is proportional to the number of pairs at that grid value. Coherent answers lie on the line $s_k+n_k=1$, and the band marks deviations of at most 0.1. Pairs in the shaded lower-left quadrant answer ``no'' to both ``the label is $X_k$'' and ``the label is not $X_k$'', and pairs in the upper-right quadrant answer ``yes'' to both. MAD is the item-averaged $|s_k+n_k-1|$ with its 95\% bootstrap CI. All three panels show the same 480 pairs.}
\label{fig:scatter}
\end{figure}

Figure~\ref{fig:scatter} plots every negation pair, and Table~\ref{tab:neg} summarizes them. \jev{}'s pairs lie close to the line $s_k+n_k=1$: the mean violation is 0.064 (95\% CI 0.055 to 0.072), and 19.0\% of its pairs deviate by more than 0.1. \qwen{}'s first-token pairs scatter widely, with a mean violation of 0.293 (0.270 to 0.316) and 66.5\% of pairs beyond 0.1. The difference, $-0.229$ ($-0.253$ to $-0.206$), holds in both domains, and none of 10{,}000 sign-flip permutations reached it ($p\approx 10^{-4}$, the smallest value the test can return). \qwen{}'s verbalized readout violates the identity by 0.122 (0.098 to 0.148), less than half as much as its first-token readout but about twice as much as \jev{}, a difference of $-0.058$ ($-0.083$ to $-0.035$). It deviates by more than 0.1 less often than \jev{} (11.9\% against 19.0\% of pairs), yet it answers a statement and its negation on the same side of 0.5 in 57 pairs, against 32 for \jev{}. Its violations are rarer than \jev{}'s but large when they occur.

\begin{table}[t]
\caption{Negation complementarity. Item means with 95\% bootstrap CIs (160 items, 480 pairs). The same-side rate is the share of pairs answered on the same side of 0.5.}
\label{tab:neg}
\centering\footnotesize\setlength{\tabcolsep}{4pt}
\providecommand{\ci}[2]{\begin{tabular}[t]{@{}c@{}}#1\\[-1.5pt]{\scriptsize$[#2]$}\end{tabular}}
\begin{tabular}{@{}lccccc@{}}
\toprule
 & \multicolumn{3}{c}{MAD, mean $|s_k+n_k-1|$} & Signed deviation & Same-side \\
\cmidrule(lr){2-4}
System & All (160 items) & NLI (100) & PubMedQA (60) & mean $(s_k+n_k-1)$ & rate \\
\midrule
Jev (native) & \ci{$0.064$}{0.055, 0.072} & \ci{$0.080$}{0.068, 0.092} & \ci{$0.036$}{0.026, 0.047} & \ci{$+0.054$}{0.045, 0.063} & \ci{$0.067$}{0.046, 0.090} \\
Qwen3.8-27B (first-token) & \ci{$0.293$}{0.270, 0.316} & \ci{$0.306$}{0.279, 0.334} & \ci{$0.270$}{0.230, 0.310} & \ci{$-0.264$}{-0.291, -0.238} & \ci{$0.417$}{0.367, 0.465} \\
\midrule
Difference (Jev $-$ Qwen) & \ci{$-0.229$}{-0.253, -0.206} & \ci{$-0.226$}{-0.257, -0.195} & \ci{$-0.234$}{-0.272, -0.196} & \ci{$+0.318$}{0.290, 0.346} & \ci{$-0.350$}{-0.402, -0.298} \\
\midrule
Qwen3.8-27B (verbalized) & \ci{$0.122$}{0.098, 0.148} & -- & -- & \ci{$+0.054$}{0.029, 0.081} & \ci{$0.119$}{0.092, 0.148} \\
\bottomrule
\end{tabular}

\end{table}

The signed deviations point in different directions. \jev{}'s mean $s_k+n_k-1$ is $+0.054$, and most of its violation is systematic, since its mean absolute violation is only slightly larger. \qwen{}'s first-token deviation is $-0.264$, and its verbalized deviation is $+0.054$, the same as \jev{}'s, so the two readouts of the same model violate the identity in opposite directions. Section~\ref{sec:forms} traces these shifts to specific question forms. In pair counts, \jev{} answers ``yes'' to both questions of a pair in 31 of 480 pairs and ``no'' to both in one. The first-token readout answers ``no'' to both in 196 pairs and ``yes'' to both in 4, and the verbalized readout answers ``yes'' to both in 47 pairs and ``no'' to both in 10. The verbalized readout is also coarse: its 960 answers take eight distinct values, 96.5\% of them take one of five, and 0.95 and 0.05 account for 70.1\%.

\jev{} is not coherent. Its two repeats of identical requests differ by 0.012 per answer on average and are identical in 39.1\% of cases, which gives a noise floor of $\nf=0.013$ (Eq.~\ref{eq:nf}). The lower confidence bound of its violation, 0.055, is more than three times this floor, so the violation is not an artifact of repeat noise. Both systems violate negation complementarity, \jev{} the least. \qwen{}'s first-token readout was not fully deterministic either: on the 20 repeated items its answers differ by 0.005 on average and are identical in 75.8\% of cases, a noise floor of 0.006, far below its violation of 0.293.

\subsection{Robustness of the difference}\label{sec:robust}

Table~\ref{tab:robust} repeats the comparison under each check. On the 253 pairs from 140 items where both systems give similar probabilities to $S_k$, \jev{}'s violation falls to 0.028 while \qwen{}'s stays at 0.257, so the gap, $-0.229$ ($-0.264$ to $-0.195$), is as large as in the full sample. The more confident system is not simply the more coherent one. Without the two double-negation labels, \jev{}'s violation is 0.057 and \qwen{}'s rises to 0.385: the pairs that remain are the ones on which \qwen{} is least coherent (Section~\ref{sec:where}), so a misreading of double negatives does not explain the gap. Averaging \jev{}'s two repeats leaves its violation at 0.063, and replacing the first repeat by the second gives 0.065. The remaining checks change the difference by at most 0.016: restricting both systems to mid-range probabilities, dropping low-mass first-token readouts, using unrounded probabilities, and splitting PubMedQA by membership in the Jevals suite. The in-suite items give about the same difference as the others ($-0.218$ and $-0.241$), so the items \jev{} may have seen during evaluation do not drive it, although this subset has only 19 items.

\begin{table}[t]
\caption{Robustness of the difference in negation violation, $\mad_{\text{neg}}$(\jev{}) $-$ $\mad_{\text{neg}}$(\qwen{}), for \qwen{}'s first-token readout unless stated. Confidence-matched pairs satisfy $|s^{J}_k-s^{Q}_k|\le 0.10$; choice-matched pairs $|c^{J}_k-c^{Q}_k|\le 0.10$. Matched subsets are averaged within items first. The double-negation labels are \emph{contradiction} and \emph{negative}. ``Mass'' is the probability the first-token readout assigns to the answer labels before renormalization, and a pair is dropped when either of its readouts falls below the bound. The logit row reports $|\mathrm{logit}\,s_k+\mathrm{logit}\,n_k|$. The first block gives the main comparison and the checks for the four confounds of Section~\ref{sec:battery}.}
\label{tab:robust}
\centering\footnotesize
\providecommand{\ci}[2]{\begin{tabular}[t]{@{}c@{}}#1\\[-1.5pt]{\scriptsize$[#2]$}\end{tabular}}
\begin{tabular}{@{}lcccc@{}}
\toprule
Comparison & Items (pairs) & Jev & Qwen & Difference [95\% CI] \\
\midrule
\multicolumn{5}{@{}l}{\emph{Main comparison and confound checks}} \\
Main comparison & 160 (480) & $0.064$ & $0.293$ & $-0.229$ {\scriptsize$[-0.253, -0.206]$} \\
Jev repeats averaged & 160 (480) & $0.063$ & $0.293$ & $-0.229$ {\scriptsize$[-0.253, -0.206]$} \\
Confidence-matched pairs & 140 (253) & $0.028$ & $0.257$ & $-0.229$ {\scriptsize$[-0.264, -0.195]$} \\
Qwen verbalized readout & 160 (480) & $0.064$ & $0.122$ & $-0.058$ {\scriptsize$[-0.083, -0.035]$} \\
Double-negation labels removed & 160 (320) & $0.057$ & $0.385$ & $-0.328$ {\scriptsize$[-0.360, -0.297]$} \\
\midrule
\multicolumn{5}{@{}l}{\emph{Further checks}} \\
Choice-matched pairs & 148 (316) & $0.033$ & $0.254$ & $-0.220$ {\scriptsize$[-0.251, -0.188]$} \\
Both $s_k\in[0.1, 0.9]$ & 114 (191) & $0.099$ & $0.332$ & $-0.232$ {\scriptsize$[-0.272, -0.194]$} \\
Both $s_k\in[0.05, 0.95]$ & 144 (259) & $0.087$ & $0.300$ & $-0.213$ {\scriptsize$[-0.246, -0.180]$} \\
Qwen probabilities not rounded & 160 (480) & $0.064$ & $0.292$ & $-0.229$ {\scriptsize$[-0.253, -0.206]$} \\
Qwen readouts with mass $\ge 0.5$ & 160 (480) & $0.064$ & $0.293$ & $-0.229$ {\scriptsize$[-0.253, -0.206]$} \\
Qwen readouts with mass $\ge 0.8$ & 160 (465) & $0.064$ & $0.292$ & $-0.228$ {\scriptsize$[-0.252, -0.205]$} \\
Jev second repeat instead of first & 160 (480) & $0.065$ & $0.293$ & $-0.228$ {\scriptsize$[-0.252, -0.205]$} \\
Logit scale & 160 (480) & $0.41$ & $2.15$ & $-1.74$ {\scriptsize$[-1.90, -1.59]$} \\
PubMedQA items in the Jevals suite & 19 (57) & $0.027$ & $0.245$ & $-0.218$ {\scriptsize$[-0.272, -0.161]$} \\
PubMedQA items outside the suite & 41 (123) & $0.040$ & $0.281$ & $-0.241$ {\scriptsize$[-0.292, -0.190]$} \\
\bottomrule
\end{tabular}

\end{table}

On logit scale the violations are 0.41 and 2.15. This check depends on the clipping bound and here works against \qwen{} rather than against the more confident system: after rounding, 18.1\% of \qwen{}'s answers to $S_k$ and $N_k$ sit at the bound of 0.01 or 0.99, against 0.5\% of \jev{}'s, although \jev{}'s answers are on average further from 0.5 (0.337 against 0.317). With a bound of 0.05 the difference shrinks from $-1.74$ to $-1.24$ ($-1.37$ to $-1.12$) and stays far from zero (exploratory). The confidence-matched comparison is therefore the stronger guard against a confidence confound.

\qwen{}'s readout quality does not explain the gap either. The median mass of its first-token readout is 0.968 (0.979 for $S_k$ questions, 0.974 for $N_k$), and keeping only readouts with mass at least 0.8 changes the difference from $-0.229$ to $-0.228$. As a check that coherence did not come at the cost of informativeness, the label with the highest $s_k$ matches the reference label in 63.8\% of items for \jev{} and 56.3\% for \qwen{}.

\subsection{Other identities}\label{sec:beyond}

Table~\ref{tab:ident} reports the other identities of Eq.~\eqref{eq:identities}. \jev{} violates most of them less than \qwen{}'s first-token readout. The exceptions are disjunction additivity, where the two are close (0.138 against 0.158), and agreement between yes/no and choice probabilities after normalization, where \qwen{} is no worse (0.085 against 0.096). Several of \jev{}'s violations are larger than its negation violation, so good negation complementarity does not imply coherence in general.

\begin{table}[t]
\caption{Other coherence identities (160 items). Item means with 95\% bootstrap CIs. The difference is \jev{} minus \qwen{}. For monotonicity, $q$ is the consequent probability ($d_k$ or $n_k$, $m\neq k$), and the share counts checks violated by more than 0.05.}
\label{tab:ident}
\centering\footnotesize
\providecommand{\ci}[2]{\begin{tabular}[t]{@{}c@{}}#1\\[-1.5pt]{\scriptsize$[#2]$}\end{tabular}}
\begin{tabular}{@{}llccc@{}}
\toprule
Identity & Statistic & Jev & Qwen (first-token) & Difference \\
\midrule
Negation & $|s_k+n_k-1|$ & $0.064$ {\scriptsize$[0.055, 0.072]$} & $0.293$ {\scriptsize$[0.270, 0.316]$} & $-0.229$ {\scriptsize$[-0.253, -0.206]$} \\
 & $s_k+n_k-1$ & $+0.054$ {\scriptsize$[0.045, 0.063]$} & $-0.264$ {\scriptsize$[-0.291, -0.238]$} & $+0.318$ {\scriptsize$[0.290, 0.346]$} \\
\addlinespace[2pt]
Disjunctive complement & $|s_k+d_k-1|$ & $0.054$ {\scriptsize$[0.048, 0.061]$} & $0.343$ {\scriptsize$[0.319, 0.369]$} & $-0.289$ {\scriptsize$[-0.315, -0.264]$} \\
 & $s_k+d_k-1$ & $+0.004$ {\scriptsize$[-0.004, 0.012]$} & $-0.327$ {\scriptsize$[-0.356, -0.299]$} & $+0.331$ {\scriptsize$[0.302, 0.360]$} \\
\addlinespace[2pt]
Partition additivity & $|\sum_k s_k-1|$ & $0.141$ {\scriptsize$[0.123, 0.159]$} & $0.319$ {\scriptsize$[0.288, 0.352]$} & $-0.178$ {\scriptsize$[-0.214, -0.144]$} \\
 & $\sum_k s_k-1$ & $+0.141$ {\scriptsize$[0.123, 0.159]$} & $-0.286$ {\scriptsize$[-0.324, -0.250]$} & $+0.427$ {\scriptsize$[0.384, 0.470]$} \\
\addlinespace[2pt]
Disjunction additivity & $|d_k-s_i-s_j|$ & $0.138$ {\scriptsize$[0.126, 0.150]$} & $0.158$ {\scriptsize$[0.146, 0.170]$} & $-0.019$ {\scriptsize$[-0.037, -0.002]$} \\
 & $d_k-s_i-s_j$ & $-0.137$ {\scriptsize$[-0.150, -0.125]$} & $-0.041$ {\scriptsize$[-0.061, -0.021]$} & $-0.096$ {\scriptsize$[-0.121, -0.071]$} \\
\addlinespace[2pt]
Unpacking & $|d_k-n_k|$ & $0.060$ {\scriptsize$[0.053, 0.067]$} & $0.106$ {\scriptsize$[0.098, 0.115]$} & $-0.046$ {\scriptsize$[-0.055, -0.038]$} \\
 & $d_k-n_k$ & $-0.050$ {\scriptsize$[-0.057, -0.044]$} & $-0.063$ {\scriptsize$[-0.073, -0.053]$} & $+0.013$ {\scriptsize$[0.003, 0.022]$} \\
\addlinespace[2pt]
Monotonicity & $\max(0, s_m-q)$ & $0.013$ {\scriptsize$[0.011, 0.016]$} & $0.033$ {\scriptsize$[0.028, 0.039]$} & $-0.020$ {\scriptsize$[-0.026, -0.014]$} \\
 & share $>0.05$ & $8.4$\% {\scriptsize$[6.8, 10.0]$} & $14.5$\% {\scriptsize$[12.9, 16.1]$} & $-6.1$\% {\scriptsize$[-8.4, -3.9]$} \\
\addlinespace[2pt]
Cross-format & $|s_k-c_k|$ & $0.091$ {\scriptsize$[0.085, 0.098]$} & $0.131$ {\scriptsize$[0.120, 0.142]$} & $-0.040$ {\scriptsize$[-0.050, -0.030]$} \\
 & $|s_k/\sum_j s_j-c_k|$ & $0.096$ {\scriptsize$[0.089, 0.102]$} & $0.085$ {\scriptsize$[0.074, 0.097]$} & $+0.010$ {\scriptsize$[-0.003, 0.023]$} \\
\bottomrule
\end{tabular}

\end{table}

\paragraph{Additivity over the partition.}
Asked separately whether each of the three labels applies, \jev{} gives probabilities that sum to 1.14 on average and to more than one on 159 of 160 items. Complementarity for two hypotheses holds much better (0.064) than additivity over three (0.141), the pattern support theory describes for human judgment~\citep{tversky1994support}. Section~\ref{sec:forms} shows that both numbers reflect the same over-endorsement of single-label statements, counted once in a negation pair and three times in a partition. \qwen{}'s first-token probabilities sum to 0.71 on average and to less than one on 132 items, the opposite of the human pattern, although its partition error (0.319) is still slightly larger than its negation error (0.293). Both systems give an explicit disjunction less probability than the sum of its parts, $d_k<s_i+s_j$, \jev{} by 0.137 and \qwen{} by 0.041.

\paragraph{Negation versus explicit disjunction.}
Support theory also predicts that describing the complement of $X_k$ as an explicit disjunction, ``$X_i$ or $X_j$'', raises its judged probability relative to the implicit description ``not $X_k$''~\citep{tversky1994support,rottenstreich1997unpacking}, that is, $d_k>n_k$. Both systems show the reverse: $d_k-n_k$ is $-0.050$ for \jev{} and $-0.063$ for \qwen{}. The two wordings of the complement differ by 0.060 on average for \jev{} and 0.106 for \qwen{}, and for \jev{} the complement phrased as a disjunction violates complementarity about as much as the negated one (0.054 against 0.064) but with a signed deviation near zero ($+0.004$, CI $-0.004$ to 0.012).

\paragraph{Monotonicity, formats, and bundling.}
Monotonicity violations are small for both systems. Label $X_m$ implies the complement of $X_k$ for $m\neq k$, so its probability should not exceed that of $D_k$ or $N_k$. The mean shortfall $\max(0,s_m-q)$ is 0.013 for \jev{} and exceeds 0.05 in 8.4\% of checks, against 0.033 and 14.5\% for \qwen{}. The yes/no and choice formats disagree for the same label by 0.091 for \jev{} and 0.131 for \qwen{}. Rescaling the yes/no probabilities to sum to one leaves a disagreement of 0.096 for \jev{} and 0.085 for \qwen{}, so for both systems the two formats carry different information rather than differing by a normalization. \jev{}'s choice probabilities always sum to one. Asking all ten questions in one request changes \jev{}'s answers no more than repeating a request does: its negation violation is 0.064 in bundled and single requests (difference $+0.0004$, CI $-0.0021$ to $0.0029$), and bundling shifts its answers by 0.0116 on average, against 0.0118 between repeats. Many of \qwen{}'s disjunction readouts put little mass on the answer tokens (29.8\% below 0.8). Keeping only readouts with mass at least 0.8, its disjunctive-complement violation is 0.278 on 335 pairs instead of 0.343 (exploratory), still far above \jev{}'s.

\subsection{Where the systematic deviations come from}\label{sec:forms}

Because exactly one of the three labels applies, any coherent assignment gives the three single-label questions of an item an average probability of 1/3 and the complement questions, negated or disjunctive, an average of 2/3, whatever the gold label is. These values are also the base rates of the events over our items, so a departure from them is at once an incoherence and a calibration-in-the-large error~\citep{dawid1982calibrated}, and it can be measured without labels. Table~\ref{tab:forms} reports the departures by question form. This decomposition is exploratory.

\begin{table}[t]
\caption{Decomposition by question form (exploratory). Mean probability of ``yes'' per question form minus the value that any coherent assignment implies when exactly one of three labels applies (Target), with 95\% bootstrap CIs. The verbalized readout was asked only the $S_k$ and $N_k$ questions.}
\label{tab:forms}
\centering\footnotesize\setlength{\tabcolsep}{4pt}
\providecommand{\ci}[2]{\begin{tabular}[t]{@{}c@{}}#1\\[-1.5pt]{\scriptsize$[#2]$}\end{tabular}}
\begin{tabular}{@{}llcccc@{}}
\toprule
System & Question form & Target & All & NLI & PubMedQA \\
\midrule
Jev (native) & $S_k$ (is $X_k$) & $1/3$ & $+0.047$ {\scriptsize$[0.041, 0.053]$} & $+0.045$ {\scriptsize$[0.038, 0.052]$} & $+0.050$ {\scriptsize$[0.040, 0.062]$} \\
 & $N_k$ (is not $X_k$) & $2/3$ & $+0.007$ {\scriptsize$[0.002, 0.013]$} & $+0.029$ {\scriptsize$[0.021, 0.037]$} & $-0.029$ {\scriptsize$[-0.036, -0.023]$} \\
 & $D_k$ (is $X_i$ or $X_j$) & $2/3$ & $-0.043$ {\scriptsize$[-0.048, -0.039]$} & $-0.043$ {\scriptsize$[-0.049, -0.038]$} & $-0.043$ {\scriptsize$[-0.050, -0.037]$} \\
\addlinespace[2pt]
Qwen (first-token) & $S_k$ (is $X_k$) & $1/3$ & $-0.095$ {\scriptsize$[-0.108, -0.083]$} & $-0.117$ {\scriptsize$[-0.132, -0.102]$} & $-0.059$ {\scriptsize$[-0.081, -0.037]$} \\
 & $N_k$ (is not $X_k$) & $2/3$ & $-0.169$ {\scriptsize$[-0.185, -0.153]$} & $-0.173$ {\scriptsize$[-0.190, -0.156]$} & $-0.162$ {\scriptsize$[-0.194, -0.130]$} \\
 & $D_k$ (is $X_i$ or $X_j$) & $2/3$ & $-0.232$ {\scriptsize$[-0.251, -0.213]$} & $-0.271$ {\scriptsize$[-0.294, -0.249]$} & $-0.165$ {\scriptsize$[-0.200, -0.131]$} \\
\addlinespace[2pt]
Qwen (verbalized) & $S_k$ (is $X_k$) & $1/3$ & $+0.009$ {\scriptsize$[-0.009, 0.026]$} & $-0.030$ {\scriptsize$[-0.050, -0.010]$} & $+0.072$ {\scriptsize$[0.042, 0.103]$} \\
 & $N_k$ (is not $X_k$) & $2/3$ & $+0.046$ {\scriptsize$[0.024, 0.069]$} & $+0.056$ {\scriptsize$[0.032, 0.083]$} & $+0.029$ {\scriptsize$[-0.012, 0.074]$} \\
\bottomrule
\end{tabular}

\end{table}

For \jev{}, single-label statements are over-endorsed by 0.047, an average probability of 0.380 for a statement that is true one time in three. Disjunctions are under-endorsed by 0.043, and negated statements are close to their target (+0.007). The signed negation deviation of $+0.054$ is the sum of the $S_k$ and $N_k$ departures, so most of it comes from the single-label questions rather than from the negated wording, and the partition excess of 0.141 is the same single-label departure summed over three labels. The disjunctive complement shows almost no signed deviation because the $S_k$ and $D_k$ departures cancel. A small yes-lean of the negated questions themselves appears only in NLI ($+0.029$) and reverses in PubMedQA ($-0.029$). \qwen{}'s first-token readout under-endorses all three forms, and the complements most: $-0.169$ with ``not'' and $-0.232$ as a disjunction, against $-0.095$ for single labels. Its deficit therefore concerns mainly complement statements rather than the word ``not'', although for \emph{affirmative} the negated form is under-endorsed more than the disjunctive one (Appendix Table~\ref{tab:bylabel}). The verbalized readout gives single-label statements about their coherent probability ($+0.009$) and over-endorses negated statements ($+0.046$). Its pairs therefore lean toward ``yes'' by about as much as \jev{}'s, but through the negated wording, whereas \jev{}'s lean comes from the single-label statements.

These departures show what single-form evaluations can and cannot see. A calibration report on \jev{}'s yes/no answers to single-label questions would find the excess, because it is a calibration-in-the-large error, but a report on choice questions would not, because \jev{}'s choice probabilities are normalized. The coherence checks find it without labels, and they also expose item-level scatter, which averages out in calibration: \qwen{}'s mean absolute negation violation exceeds the magnitude of its mean signed deviation, 0.293 against 0.264 for the first-token readout and 0.122 against 0.054 for the verbalized one, so its violations are not a constant shift, least of all in the verbalized readout.

\subsection{Where the violations occur}\label{sec:where}

The two systems agree on which labels are likely: the Spearman correlation of their $s_k$ is 0.87 (0.84 to 0.89; bootstrap over pairs). They overlap little on where coherence fails: the correlation of their violation magnitudes $|s_k+n_k-1|$ is 0.18 (0.07 to 0.27; exploratory, bootstrap over items).

\begin{figure}[t]
\centering
\includegraphics[width=\linewidth]{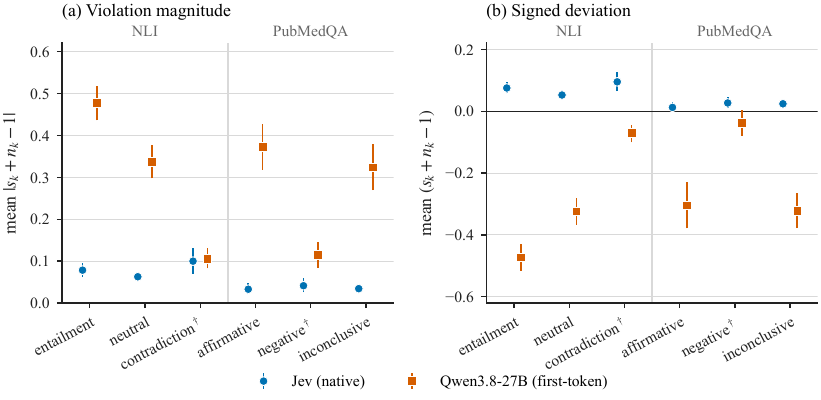}
\caption{Negation violations by label, with 95\% bootstrap CIs (100 NLI and 60 PubMedQA items). $^\dagger$Labels that denote falsity, whose negation is a double negative. (a) Mean $|s_k+n_k-1|$. (b) Mean $s_k+n_k-1$, where negative values mean that the two answers of a pair together give too little probability to ``yes''.}
\label{fig:label}
\end{figure}

Figure~\ref{fig:label} breaks the violations down by label. \qwen{}'s first-token violations are largest for \emph{entailment} (0.477) and \emph{affirmative} (0.372), with signed deviations of $-0.472$ and $-0.305$. For \emph{entailment}, \qwen{} gives on average 0.11 to ``the label is entailment'' and only 0.42 to ``the label is not entailment''. Its violations are smallest for the two double-negation labels (0.105 and 0.114). \jev{}'s largest violation is on \emph{contradiction} (0.100), and its signed deviations are positive for all six labels. Appendix Table~\ref{tab:bylabel} gives the numbers.

\begin{figure}[t]
\centering
\includegraphics[width=\linewidth]{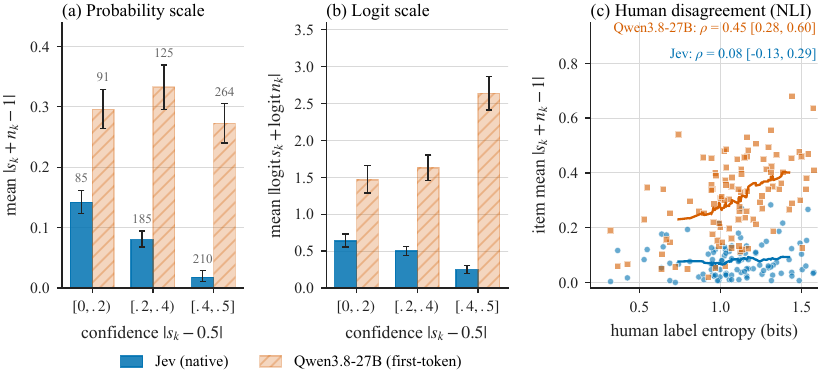}
\caption{(a, b) Mean negation violation by the confidence of the answer to $S_k$, $|s_k-0.5|$, on the probability and logit scales, with the number of pairs above the bars in (a) and 95\% intervals from a bootstrap over items. (c) Item-level violation against the entropy of 100 human labels on ChaosNLI. Lines are running means over 25 items, and $\rho$ is the Spearman correlation with its 95\% bootstrap CI.}
\label{fig:conf}
\end{figure}

Figure~\ref{fig:conf} relates the violations to confidence. \jev{}'s violations shrink as its answer to $S_k$ becomes more confident, from 0.142 (0.124 to 0.162) for pairs with $|s_k-0.5|<0.2$ to 0.018 (0.010 to 0.029) for pairs with $|s_k-0.5|\ge 0.4$. The probability scale leaves little room for deviation in the most confident group, but the drop remains on logit scale, from 0.64 to 0.24. \jev{}'s yes-lean is concentrated in the same place: its signed deviation falls from $+0.13$ (0.11 to 0.15) in the least confident group to $+0.01$ (0.00 to 0.02) in the most confident, and 22 of its 31 ``yes to both'' pairs fall in the least confident group. \qwen{}'s violations do not shrink with confidence. On probability scale they are 0.296, 0.333, and 0.272 across the three groups, and on logit scale they are largest on the 264 pairs where it is most confident (2.64). Many of these pairs have $s_k$ near 0 and $n_k$ well below 1 (Figure~\ref{fig:scatter}b): the readout is confident that the label is not $X_k$ and still does not endorse the statement that the label is not $X_k$. Its rate of answering ``no'' to both is highest where it is least confident, 60\%, 58\%, and 26\% of pairs across the three groups.

On the NLI items, \qwen{}'s item-level violation increases with the entropy of the human labels ($\rho=0.45$, 0.28 to 0.60). For \jev{} the correlation is 0.08 ($-0.13$ to 0.29), and we detect no association. The two correlations differ by 0.37 (0.10 to 0.63; exploratory, bootstrap over items).

\section{Discussion}\label{sec:discussion}

\paragraph{Coherence and calibration.}
Coherence and calibration overlap but do not coincide. When a partition is known, the systematic part of an incoherence is also a calibration-in-the-large error, as \jev{}'s over-endorsement of single-label statements shows. Coherence checks find such biases without labels, attribute them to question forms, and add the item-level inconsistencies that calibration averages away. \jev{} gives a statement and its negation nearly complementary probabilities, with residual violations that are small and concentrated on uncertain cases. Its larger violations appear when more than two outcomes are involved, because the single-label excess accumulates over a partition: three separate yes/no questions overcount it by 0.14 on average, and yes/no and choice probabilities for the same label differ by 0.09. In practice, probabilities from different question forms should not be mixed in one decision rule, a set of per-option yes/no probabilities should not be treated as a distribution, and the choice question is coherent by construction for a partition. The developer's own guidance, not to carry a threshold tuned on one format over to another~\citep{typesafe2026jaggedness}, is consistent with these measurements.

\paragraph{What LLM baselines measure.}
The two readouts of \qwen{} differ in size and direction. The first-token readout under-endorses complements and rejects both a statement and its negation in 196 of 480 pairs. The verbalized readout leans toward accepting both and violates the identity less than half as much (0.122 against 0.293). The incoherence we observe is therefore partly a property of the readout rather than of the model's underlying judgment, in line with evidence that first-token probabilities and written answers diverge~\citep{wang2024myanswer}. Benchmarks that compare typed-decision models with LLM baselines through a single readout, as current evaluations do~\citep{jevalsmethod2026}, measure a combination of model and elicitation method. A coherence battery makes this visible because it asks the same system logically equivalent questions.

\paragraph{Relation to earlier audits.}
Our results agree in direction with the community audits of \jev{}: Jevify's aggregate sure loss is lower for \jev{} than for untuned open-LLM readouts~\citep{jevify2026}, and we find the same ordering for negation complementarity once confidence, readout, and double negation are controlled. The disagreement between yes/no and choice probabilities, 0.09 in our data, lies between the 0.055 and 0.125 reported for judging tasks and two-option choices~\citep{li2026jevjudge,jujumilk2026audit}. Our study has no paraphrase control, and one audit found negation violations no larger than those produced by rewording a question~\citep{yodablocks2026orderby}. Within our battery, the two wordings of the complement differ by 0.060 on average, about as much as the negation violation itself, so part of what we measure may be sensitivity to wording rather than to the logical relation.

\paragraph{Mechanism.}
\jev{}'s architecture is not public, and its developer describes the training only as reinforcement learning for calibrated decisions~\citep{typesafe2026jev}, so we cannot attribute its coherence to a mechanism. The observations constrain explanations. \jev{}'s systematic departures are tied to question forms, over-endorsing single labels and under-endorsing explicit disjunctions, and its residual violations shrink with confidence on both scales and do not track human disagreement. \qwen{}'s first-token readout under-endorses complement statements in either wording, and its violations do not shrink where it is confident about the label, which points to how the readout handles complement questions rather than to uncertainty about the label. \jev{}'s yes/no probabilities also never reached 0 or 1: all 4{,}320 of its yes/no answers, across both repeats and the bundled requests, lie in $[0.01,0.99]$, in line with the two-decimal output grid documented at larger scale~\citep{rafe2026crash}.

\section{Limitations}\label{sec:limits}

The study is small by design: 160 items, 480 negation pairs, two English domains with three-label partitions, and one wording each for negation and disjunction, without a paraphrase control (Section~\ref{sec:discussion}). The confidence intervals are narrow for the main comparison, but per-label and subset estimates rest on 19 to 100 items. The comparison LLM is a single model, run with thinking disabled. Larger models, other model families, or reasoning modes may be more coherent, as GPT-5's near-perfect binary complementarity suggests~\citep{imannezhad2026divergent}, so our comparison speaks to this model and these readouts rather than to LLMs in general. We tested one \jev{} version through one gateway, and a closed, versioned service can change. Both datasets are public, so memorization cannot be excluded, although the Jevals-suite items behaved like the others. Finally, coherence is necessary for useful probabilities but not sufficient: a system that answers 1/3 to every single-label question and 2/3 to every complement question satisfies every identity in our battery and conveys nothing about the item. The accuracy check in Section~\ref{sec:robust} shows that \jev{}'s answers were informative in this sample.

\section{Conclusion}\label{sec:conclusion}

A battery of logically linked questions shows that \jev{}'s native probabilities come closer to the probability axioms than two readouts of an open-weight LLM on the same questions: its violation of negation complementarity is 0.064, against 0.293 and 0.122 for the first-token and verbalized readouts of the official \qwen{} weights. The difference survives controls for confidence, readout, double negation, and repeat noise. \jev{}'s probabilities are still not coherent. It over-endorses single-label statements, which overcounts a three-way partition by 0.14 and accounts for most of its residual negation deviation, and its yes/no and choice probabilities differ. Coherence checks need no labels and cost little. Applied alongside calibration, they tell users of native probabilities which question forms they can combine.

\section*{Data and code availability}
The stimuli, raw request and response logs, analysis code, and the scripts that produce every table and figure are available at \url{https://github.com/bro789/typed-decision-coherence}.

\bibliographystyle{unsrtnat}
\bibliography{references}

\clearpage
\appendix

\section{Question text, prompts, and reproducibility details}\label{app:prompts}

\paragraph{Definitions.} Every question of an item begins with the definition paragraph of its domain.
\begin{itemize}
\item[] \textbf{NLI.} \emph{Classify the relation between the premise and the hypothesis. There are exactly three mutually exclusive labels: entailment (given the premise, the hypothesis is definitely true), neutral (given the premise, the hypothesis might or might not be true), and contradiction (given the premise, the hypothesis is definitely false).}
\item[] \textbf{PubMedQA.} \emph{Classify the conclusion that the context passages support for the research question. There are exactly three mutually exclusive labels: affirmative (the passages support answering the research question `yes'), negative (the passages support answering the research question `no'), and inconclusive (the evidence in the passages leaves the answer uncertain: `maybe').}
\end{itemize}

\paragraph{Questions.} The definition paragraph is followed by one of: ``Is it true that the label is $X_k$?''; ``Is it true that the label is not $X_k$?''; ``Is it true that the label is either $X_i$ or $X_j$?''; or ``Which label applies?''. \jev{} receives the state as the \texttt{state} field and each question as a \texttt{noul} question (the three yes/no forms, without optional criteria) or a \texttt{choice} question whose options carry the label definitions. For \qwen{}, the system prompt is ``You answer typed decision questions about a state. Base your answer on the state. Follow the answer format exactly.'' The user message is ``State:'' followed by the state, ``Question:'' followed by the question, and the instruction ``Answer with one word: yes or no.'' for yes/no questions or a lettered option list with ``Reply with the letter of the single best option only.'' for the choice question. The verbalized readout replaces the last instruction with ``Give a probability for each answer. Reply only with JSON: \{"probabilities": \{"yes": <p>, "no": <p>\}\}''.

\paragraph{Example.} NLI item \texttt{104395c}: \emph{Premise: Routine screening and intervention will require engendering a sense of role responsibility among emergency department clinicians towards addressing substance abuse. Hypothesis: Routine screening has no impact on substance abuse prevention.} The 100 human annotators chose entailment, neutral, and contradiction with frequencies 0.03, 0.37, and 0.60. One of its ten questions is the definition paragraph followed by ``Is it true that the label is not entailment?''.

\paragraph{Readout details.} The messages for \qwen{} are rendered with the model's own chat template. The first-token readout requests the 100 most probable next tokens at temperature 0, with log-probabilities taken before any sampling adjustment, strips whitespace from each token, lowercases it for yes/no questions, and sums the probability of the tokens that match an answer label. The verbalized readout generates at most 300 tokens at temperature 0 with thinking disabled, parses the text between the first opening and the last closing brace of the reply as JSON, keeps the entries for ``yes'' and ``no'', and renormalizes them. All 960 replies yielded usable probabilities. \jev{} answers that were not valid probabilities would have been re-asked once, which never occurred. The bootstrap and permutation seeds are 0. The \jev{} calls were made on 25 September 2026 and the \qwen{} calls on 26 September 2026.

\section{Results by label}\label{app:bylabel}

\begin{table}[H]
\caption{Negation violations by (domain, label). $^\dagger$Double-negation labels. The logit MAD is the mean of $|\mathrm{logit}\,s_k+\mathrm{logit}\,n_k|$ with probabilities clipped to $[0.01,0.99]$.}
\label{tab:bylabel}
\centering\footnotesize\setlength{\tabcolsep}{3pt}
\providecommand{\ci}[2]{\begin{tabular}[t]{@{}c@{}}#1\\[-1.5pt]{\scriptsize$[#2]$}\end{tabular}}
\begin{tabular}{@{}llccccccccc@{}}
\toprule
 & & \multicolumn{3}{c}{$|s_k+n_k-1|$} & \multicolumn{2}{c}{$s_k+n_k-1$} & \multicolumn{2}{c}{$d_k-n_k$} & \multicolumn{2}{c}{logit MAD} \\
\cmidrule(lr){3-5}\cmidrule(lr){6-7}\cmidrule(lr){8-9}\cmidrule(lr){10-11}
Domain & Label $X_k$ & Jev & Qwen & Difference [95\% CI] & Jev & Qwen & Jev & Qwen & Jev & Qwen \\
\midrule
NLI & entailment & $0.078$ & $0.477$ & $-0.398$ {\scriptsize$[-0.444, -0.351]$} & $+0.075$ & $-0.472$ & $-0.057$ & $-0.043$ & $0.57$ & $3.77$ \\
 & neutral & $0.063$ & $0.336$ & $-0.274$ {\scriptsize$[-0.317, -0.231]$} & $+0.052$ & $-0.324$ & $-0.022$ & $-0.159$ & $0.36$ & $1.85$ \\
 & contradiction$^\dagger$ & $0.100$ & $0.105$ & $-0.005$ {\scriptsize$[-0.037, 0.025]$} & $+0.095$ & $-0.071$ & $-0.138$ & $-0.094$ & $0.63$ & $1.26$ \\
\midrule
PubMedQA & affirmative & $0.033$ & $0.372$ & $-0.339$ {\scriptsize$[-0.392, -0.285]$} & $+0.012$ & $-0.305$ & $-0.022$ & $+0.080$ & $0.23$ & $2.94$ \\
 & negative$^\dagger$ & $0.041$ & $0.114$ & $-0.073$ {\scriptsize$[-0.107, -0.039]$} & $+0.027$ & $-0.038$ & $-0.014$ & $-0.099$ & $0.28$ & $0.80$ \\
 & inconclusive & $0.034$ & $0.324$ & $-0.289$ {\scriptsize$[-0.343, -0.237]$} & $+0.024$ & $-0.322$ & $-0.006$ & $+0.010$ & $0.19$ & $2.01$ \\
\bottomrule
\end{tabular}

\end{table}

\section{Readout quality, noise, and cost}\label{app:readout}

\begin{table}[H]
\caption{Calls, readout quality, repeat noise, and cost. During data collection the \qwen{} model server handled no other requests and its GPUs ran no other jobs. Its latency is the wall-clock time of a first-token call with four calls in flight.}
\label{tab:readout}
\centering\footnotesize
\providecommand{\ci}[2]{\begin{tabular}[t]{@{}c@{}}#1\\[-1.5pt]{\scriptsize$[#2]$}\end{tabular}}
\begin{tabular}{@{}ll@{}}
\toprule
Quantity & Value \\
\midrule
Jev calls (single-question + bundled) & 3{,}360 (3{,}360 succeeded) \\
Qwen first-token calls & 1{,}720 (1{,}720 parsed) \\
Qwen verbalized calls & 960 (960 parsed) \\
Jev total cost & \$0.0757 \\
Jev latency, median / p95 (s) & 0.64 / 0.75 \\
Qwen first-token latency, median / p95 (s) & 0.44 / 0.90 \\
Qwen label mass, median: all / S / N / D / choice & 0.968 / 0.979 / 0.974 / 0.877 / 0.995 \\
Qwen label mass $<0.8$: all / S / N / D / choice & 9.2\% / 1.5\% / 1.5\% / 29.8\% / 0.0\% \\
Jev repeat noise: mean $|p^{(1)}-p^{(0)}|$; identical share & 0.0118; 39.1\% \\
Jev noise floor $\mathrm{NF}$ [95\% CI] & 0.0127 [0.0112, 0.0142] \\
Qwen repeat noise (20 items): mean $|p^{(1)}-p^{(0)}|$; identical share & 0.0051; 75.8\% \\
Qwen noise floor $\mathrm{NF}$ [95\% CI] & 0.0065 [0.0042, 0.0087] \\
\bottomrule
\end{tabular}

\end{table}

\section{Exploratory analyses}\label{app:posthoc}

The following quantities were computed after the main analysis, from the same parsed answers, by the script that produces the tables and figures:
\begin{itemize}
\item quadrant and confidence-bin counts of pairs answered ``yes'' to both or ``no'' to both questions;
\item the same-side rate and the share of deviations above 0.1 for the verbalized readout;
\item the decomposition by question form (Table~\ref{tab:forms});
\item the correlation of violation magnitudes between the systems, and the difference between their human-disagreement correlations;
\item the logit-scale confidence bins and all confidence-bin intervals;
\item the clipping sensitivity of the logit measure and the mass-filtered disjunctive identities;
\item the counts of items whose single-label probabilities sum above or below one, and the mean probabilities for \emph{entailment};
\item the value range of \jev{}'s yes/no answers and the value distribution of the verbalized readout.
\end{itemize}
They describe the data further and are not used to test the main comparison.

\end{document}